\documentclass[10pt, a4paper]{article}

\usepackage{lrec-coling2024} 

\usepackage{natbib}
\usepackage{multibib}
\makeatletter
\def\@mb@citenamelist{cite,citep,citet,citealp,citealt,citepalias,citetalias}
\makeatother
\newcites{languageresource}{~}

\usepackage{graphicx}
\usepackage{tabularx}
\usepackage{soul}
\usepackage{titlesec}
\titleformat{\section}{\normalfont\large\bfseries\center}{\thesection.}{1em}{}
\titleformat{\subsection}{\normalfont\SmallTitleFont\bfseries\raggedright}{\thesubsection.}{1em}{}
\titleformat{\subsubsection}{\normalfont\normalsize\bfseries\raggedright}{\thesubsubsection.}{1em}{}
\renewcommand\thesection{\arabic{section}}
\renewcommand\thesubsection{\thesection.\arabic{subsection}}
\renewcommand\thesubsubsection{\thesubsection.\arabic{subsubsection}}

\usepackage{bbding}

\usepackage{xcolor}
\usepackage{hyperref}
 \definecolor{darkblue}{rgb}{0, 0, 0.5}
  \hypersetup{colorlinks=true, citecolor=darkblue, linkcolor=darkblue, urlcolor=darkblue}

\usepackage{xstring}

\usepackage{color}

\usepackage[utf8]{inputenc}

\usepackage{multirow}
\usepackage{CJKutf8}

\title{Annotations for Exploring Food Tweets From Multiple Aspects}

\name{Matīss Rikters$^\lambda$, Edison Marrese-Taylor$^\lambda$, Rinalds Vīksna$^\iota$} 

\address{
    $^\lambda$ Artificial Intelligence Research Center, \\
    National Institute of Advanced Industrial Science and Technology \\
    \{firstname.lastname\}@aist.go.jp\\ 
    $^\iota$ University of Latvia\\ 
    \{rinaldsviksna\}@gmail.com
}

\abstract{
    This research builds upon the Latvian Twitter Eater Corpus (LTEC), which is focused on the narrow domain of tweets related to food, drinks, eating and drinking. LTEC has been collected for more than 12 years and reaching almost 3 million tweets with the basic information as well as extended automatically and manually annotated metadata. In this paper we supplement the LTEC with manually annotated subsets of evaluation data for machine translation, named entity recognition, timeline-balanced sentiment analysis, and text-image relation classification. We experiment with each of the data sets using baseline models and highlight future challenges for various modelling approaches.
    \\ \newline 
    \Keywords{Social Media Analysis, Corpus Creation, Multimodality} 
}

\begin{document}

\maketitleabstract

\section{Introduction}

Despite the recently induced chaos due to the company leadership change, Twitter (now renamed to X\footnote{From Twitter to X: Elon Musk Begins Erasing an Iconic Internet Brand - \url{https://www.nytimes.com/2023/07/24/technology/twitter-x-elon-musk.html}}) has long been and still remains one of the most influential social networks not only around political or technological topics, but also for everyday lifestyle content and regular people posting about their daily lives. For years Twitter was also one of the only remaining useful social media platforms to the research community by providing real-time access to new posts with additional elevated access for academic research purposes at no cost\footnote{Guide to the future of the Twitter API - \url{https://developer.twitter.com/en/products/twitter-api/early-access/guide}}. This, however, also became deprecated in fall of 2023\footnote{\url{https://developer.twitter.com/en/updates/changelog}} with no reasonable alternative so far, aside from the \$5000.00 USD/month Pro tier.

The Latvian Twitter Eater Corpus (LTEC \cite{SprogisRikters2020BalticHLT}) is a collection of tweets gathered by following the appearance of 363 keywords related to food and eating inflected in various valid word forms in the Latvian language. Data collecting was started in 2011 and has reached over 2.5 million tweets generated by more than 170,000 users. Each tweet in the dataset is represented by standard data fields like tweet date, text and id, author screen name, as well as additional metadata fields where available, such as location details, attached media information, a list of food and drink products mentioned in the tweet text in original surface forms and nominative singular forms (with respective English translations), and a separate subset of tweets with manually annotated sentiment classes - positive, neutral and negative.

In this paper we describe supplementing the LTEC with several new manually annotated and task-specific evaluation sets for text-image relation classification, machine translation, named entity recognition and an additional set for sentiment analysis. We also provide results from baseline experiments using each of these evaluation sets

\section{Related Work}

The LTEC has been analysed from several aspects of sensory experiences related to the food products mentioned in the tweet content by \citet{kale-tracing}, as well as comparing the change of sentiment in food tweets over time. Most recently \cite{kale2023food} linked the timeline of food tweet sentiment changes with weather observation data in the area for further insights. We believe that our additions to the dataset can further spark analytical research of food-related tweet data from other aspects, such as the images related to tweets.

\citet{vempala-preotiuc-pietro-2019-categorizing} explore the relations between Tweet text and attached images and categorise them in four different groups: a) the image adds to the text meaning and the text is represented in the image; b) the image adds to the text meaning and the text is not represented in the image; c) the image does not add to the text meaning and the text is represented in the image; and d) the image does not add to the text meaning and the text is not represented in the image. They also further analyse user demographic traits linked to each of the four image tweeting types.


\section{Manually Annotated Datasets}

In this section, we discuss the newly created manually annotated parts of LTEC. For the machine translation (MT) and named entity recognition (NER) tasks, we manually translated and annotated the same test set of 744 tweets, which already had manually assigned sentiment classes. This test set, however, was far too poor in terms of attached images, so for the text-image relation task we sampled 800 recent tweets with image attachments. Finally, for the extended and more balanced sentiment analysis test set, we sampled 50 random tweets from each year between 2011 and 2020, totalling 500.

The number of annotators was task-specific, having 12 annotators for sentiment analysis, a total of 3 for NER and image-text relation, and 2 for translation. With some of them overlapping between tasks, there were a total of 16 unique annotators. A more detailed annotator profile is provided in the appendix along with annotation environment descriptions and annotation instructions.

\subsection{Text-image Relation}

\begin{figure}[t]
   \centering
   \includegraphics[width=\linewidth]{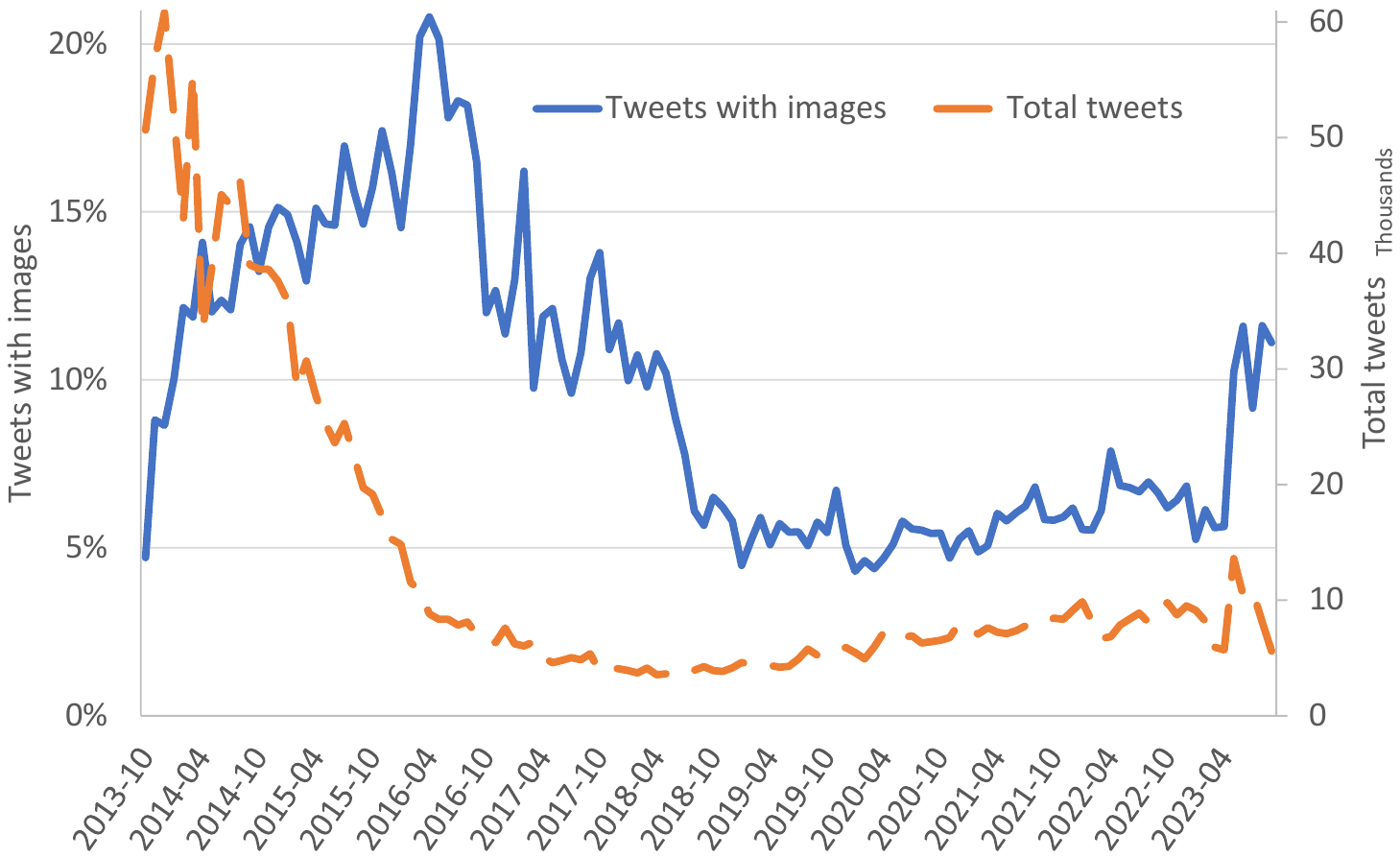}
   \caption{The total count of monthly tweets contrasted to the percentage of monthly tweets containing images.}
   \label{fig:image-tweet-couts}
\end{figure}

While not available from the very beginning of Twitter, the option to embed images directly into a tweet was added in 2011 \cite{photo-sharing} after TwitPic had been the main external source of attached images for several years. The popularity of posting images along with food-related tweets has shifted over the years between 5 and 20 per cent of total monthly tweets, as can be seen in Figure \ref{fig:image-tweet-couts}.
We use the same annotation schema as \citet{vempala-preotiuc-pietro-2019-categorizing} by grouping image-text relations into whether the image adds to the text meaning and whether the text is represented in the image. We manually annotate a test set of 800 image -- tweet pairs. In our annotations we found that the majority do indeed textually describe what is represented in the image, and about half of the cases the image also adds to the meaning of the text. Only around 9\% add to the meaning without describing the contents in the text, and a mere 6\% do neither.

\begin{figure}
    \centering
    \includegraphics[width=1\linewidth]{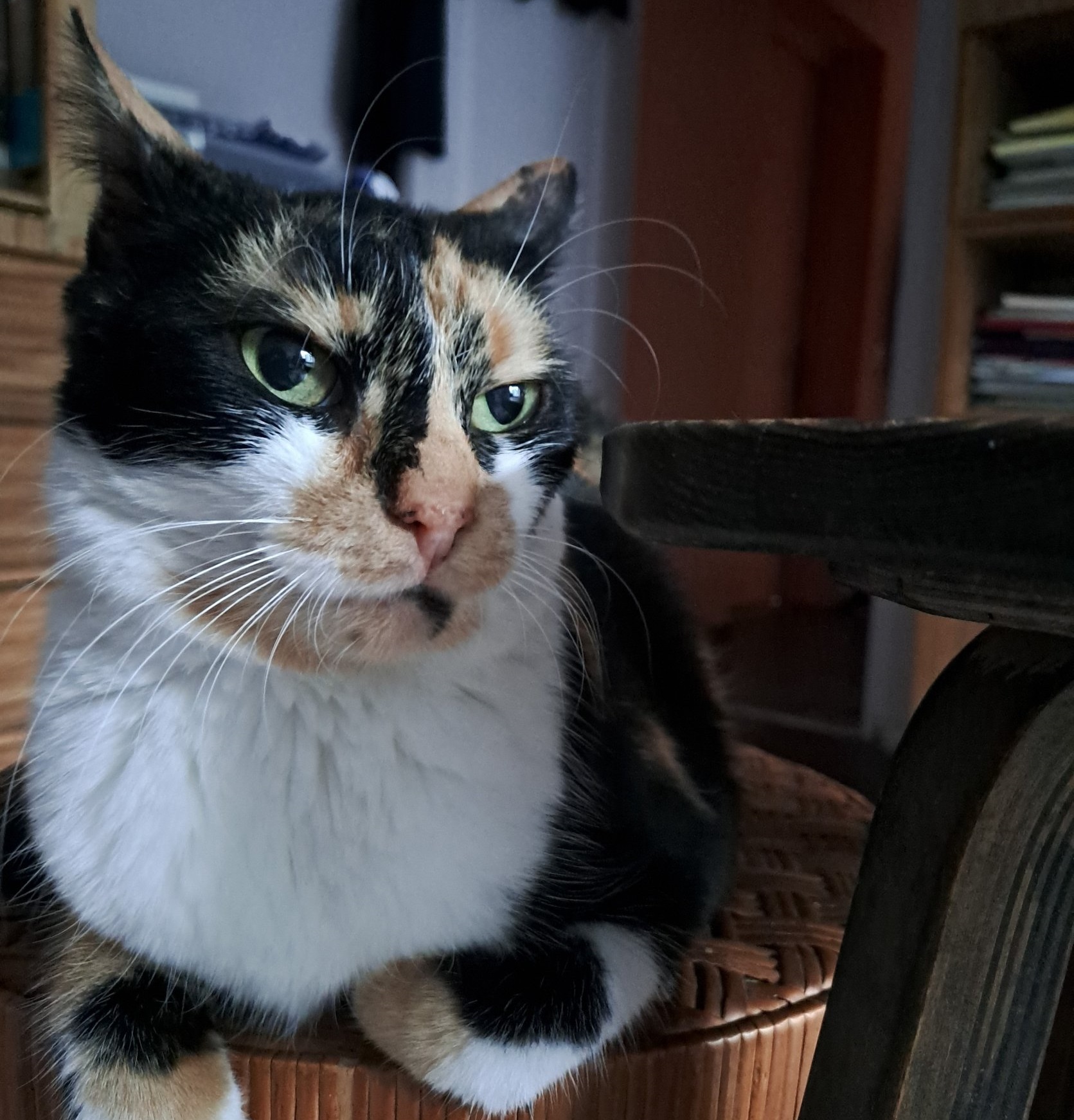}
    \begin{tabular}{l|l}
        Text & \textit{Brokastis ēd! Nu ēd, ēd!!!} \\
        Translation & \textit{Eat breakfast! Well, eat, eat!!!}
    \end{tabular}
    \caption{The attached image to the tweet text mentioned in Table \ref{tab:prompt-example}.}
    \label{fig:prompt-example-image}
\end{figure}

For example, given the image shown in Figure \ref{fig:prompt-example-image} and its accompanying text from the tweet (and its English translation), it is fairly clear that the text does not at all describe what is shown in the image, since we can see a cat sitting on a chair in the image and it is not mentioned anywhere in the text. However, the image most certainly adds to the meaning of the text -- upon seeing the angry look of the cat it becomes clear that the text represents its personified thoughts.

\subsection{Machine Translation}

Manual translation of the 744 tweet test set from the original Latvian language into English was performed by a single translator. After the translation the sentences were reviewed by a professional post-editor as an additional quality assurance measure. Such texts tend to be challenging to comprehend and convey by modern models due to the use of specific terminology and abbreviations, which at times are used to maintain posts within certain character count limitations of social networks.

\subsection{Named Entity Recognition}

The 744 tweet test set was annotated in the CoNLL-2003 format \cite{tjong-kim-sang-de-meulder-2003-introduction} with four entity classes for persons, locations, organisations and miscellaneous (PER, LOC, ORG, MISC), mostly following MUC-7 guidelines \cite{chinchor-1998-overview}. 
Annotations were performed with a reduced entity set similar to the Balanced State-of-the-Art Multilayer Corpus for natural language understanding\footnote{\url{https://github.com/LUMII-AILab/FullStack/tree/master/NamedEntities}} \cite{gruzitis-etal-2018-creation} by converting the nine entity tag set to a four entity tag set. We kept the PERSON and ORGANIZATION tags, joined the LOCATION and GPE tags into a single LOCATION tag, and joined the remaining tags into a single MISC category. The manual annotation process was carried out by 2 annotators using the INCEpTION \cite{klie-etal-2018-inception} tool, with a third annotator resolving conflicts. Cohen’s Kappa (as reported by the INCEpTION) between two main annotators for the NER task was 0.92. The test set contains 188 MISC entities, 99 LOC entities, 55 PER entities, and 68 ORG entities. We also introduced a new FOOD tag to be used for marking food and drink related entities.

\subsection{Sentiment Analysis}

The original sentiment analysis test set in LTEC was sampled from tweets of only one year while the whole corpus spans over a decade. While translations and named entities are somewhat more independent from the specific time period, sentiment can be more time-sensitive. For a more trustworthy evaluation, we decided to randomly select 50 tweets per year from each year between 2011 and 2020 into a more balanced additional 500 tweet evaluation set. Twelve human evaluators were asked to individually judge the sentiment for each of the 500 tweets. We used the majority vote of the human evaluators as the final annotation in cases where they disagreed on a particular evaluation and considered two classifications as correct in the 21 cases where the majority opinion was split in half (for example, 6 positive and 6 neutral). The overall agreement of the evaluators was 70.48\% with a free marginal kappa \citep{randolph2005free} of 0.56 (values from 0.40 to 0.75 are considered intermediate to good agreement). This shows that the tweet texts are not always trivial enough to be unequivocally classified into just one of the three sentiment classes.

\section{Experiments}

\subsection{Text-image Relation}

Many modern multi-modal large language models have recently gained the ability to answer questions about a given image in plain text chat form. In the text-image relation experiments we feed our data to the LLaVA \cite{liu2023llava} model\footnote{llava-v1-0719-336px-lora-merge-vicuna-13b-v1.3} and prompt it to answer a question about the type of text-image relation using the two prompts outlined in Table \ref{tab:prompt-example}. We experimented with providing the model with the original tweet texts written in Latvian, as well as automatically translated versions of tweet texts into English using Tilde MT\footnote{Tilde MT October 2023 version - \url{https://translate.tilde.com/}}, since the LLaVA model is using Vicuna \cite{zheng2023judging} as its language model part, which is inherently not multilingual. Prediction accuracy was only 20.69\% when evaluated on the original texts, and improved slightly to 27.83\% on English translations. The results in Table \ref{tab:image-prediction} show more detailed results, which outline how the model tends to mostly answer the first question positively at around 84-88\% of the time, while it should be around 57\%. Meanwhile, the model seems to be answering the second question overly negatively at around 56-67\%, where the answer should be closer to 15\%, since in the vast majority of annotated tweets the content of the text was indeed represented in the image.

\begin{table}
    \begin{tabular}{l|l}
        Prompt 1 & Given the following text, extracted \\
        & from a tweet in Latvian: \\
        & \textit{Brokastis ēd! Nu ēd, ēd!!!} \\
        & Is the image adding to the text \\
        & meaning? Reply ``Yes" or ``No". \\ \hline 
        Prompt 2 & Given the following text, extracted \\ 
        & from a tweet in Latvian: \\
        & \textit{Brokastis ēd! Nu ēd, ēd!!!} \\
        & Is the text represented in the image? \\ 
        & Reply ``Yes" or ``No".
    \end{tabular}
    \caption{Examples of the prompts used for querying the LLaVA model.}
    \label{tab:prompt-example}
\end{table}

\begin{table}
    \centering
    \begin{tabular}{c|ccc}
         & Gold & Predicted & Correctly\\
         & \multicolumn{3}{c}{Latvian} \\ \hline 
        IA \Checkmark \ TR \Checkmark \ & 392 & 240 & 113 \\
        IA \Checkmark \ TR \XSolidBrush \ & 72 & 442 & \textbf{35} \\
        IA \XSolidBrush \ TR \Checkmark \ & 296 & 31 & \textbf{8} \\
        IA \XSolidBrush \ TR \XSolidBrush \ & 52 & 99 & 12 \\
         & \multicolumn{3}{c}{English} \\ \hline 
        IA \Checkmark \ TR \Checkmark \ & 392 & 336 & \textbf{172} \\
        IA \Checkmark \ TR \XSolidBrush \ & 72 & 379 & \textbf{35} \\
        IA \XSolidBrush \ TR \Checkmark \ & 296 & 20 & 6 \\
        IA \XSolidBrush \ TR \XSolidBrush \ & 52 & 77 & \textbf{13} \\
    \end{tabular}
    \caption{Automatic prediction results by LLaVA. IA represents the image adding to the text meaning, TR -- the text being represented in the image, \Checkmark \ and \XSolidBrush \ -- true or false respectively.}
    \label{tab:image-prediction}
\end{table}


\subsection{Machine Translation}

We evaluated several state of the art MT models using the dataset and compared automatic evaluation according to BLEU \cite{papineni-etal-2002-bleu} and ChrF \cite{popovic-2015-chrf} scores using sacreBLEU \cite{post-2018-call}. We compare the translations from publicly available translation services such as Google Translate\footnote{Google Translate October 2023 version \url{https://translate.google.com}} and Tilde MT\footnote{Tilde MT October 2023 version - \url{https://translate.tilde.com/}}, and open source pre-trained models like Opus MT \cite{TiedemannThottingal:EAMT2020} and mBART \cite{tang-etal-2021-multilingual}. The results summarised in Table \ref{tab:mt-result-table} indicate that such social media data is not particularly challenging to translate. While mBART does noticeably fall behind the others three, BLEU scores in the range of 40 to 50 usually indicate fairly usable translations. Upon manual inspection there was one very noticeable frequent error made by the pre-trained models which was absent in the output from translation services and is directly related to typical modern social media texts -- partial or full absence of emoji and emoticons.

\begin{table}
    \centering
    \begin{tabular}{l|cc}
         & BLEU & ChrF \\ \hline
        Google Translate & 43.09 & 65.75 \\
        Tilde MT & \textbf{48.28} & \textbf{68.21} \\
        Opus MT & 41.73 & 62.38 \\
        mBART & 33.36 & 54.45 \\
    \end{tabular}
    \caption{Machine translation experiment results.}
    \label{tab:mt-result-table}
\end{table}

\subsection{Named Entity Recognition}

The task of named entity recognition (NER) has been proven more challenging for languages with less widely available data, and especially morphologically rich languages such as Latvian. Recognising named entities (NEs) in texts from social media adds an extra layer to the challenge, as such texts are not necessarily the most grammatically correct, and even tend to use specific terminology or abbreviations to maintain the content of the post within the limits of the specific social media network (for Twitter formally 140, later changed to 280 symbols). 

We trained 2 NER models using \cite{fullstack} dataset\footnote{\url{https://github.com/LUMII-AILab/FullStack}}: a fine-tuned multilingual BERT (mBERT \cite{devlin-etal-2019-bert}) and fine-tuned mBERT additionally pre-trained on tweets and added emoticons vocabulary\footnote{\url{https://huggingface.co/FFZG-cleopatra/bert-emoji-latvian-twitter}} \cite{thakkar-2020-sentiment} using the Flair library \cite{schweter2020flert}.

\begin{table}
    \centering
    \begin{tabular}{lcc}
    Entity & mBERT & mBERT+tweets \\
    \hline 
    LOC    & 88.44 & \textbf{94.00} \\
    MISC   & \textbf{85.25} & 80.75 \\
    ORG    & 75.00 & \textbf{83.82} \\
    PER    & 85.22 & \textbf{89.66} \\
    \hline 
    Overall& 84.41 & \textbf{85.71} \\
    \end{tabular}
    \caption{Results (F\textsubscript{1} scores) of applied NER.}
    \label{tab:ner}
\end{table}
The results in Table \ref{tab:ner} show that a language model additionally pre-trained on Latvian tweets performs better when applied to in-domain data in a downstream task. The overall F\textsubscript{1} score is lower than SOTA (88.1) \cite{LitLat} on the Fullstack \cite{fullstack} dataset, however, the LitLat score is obtained performing the NER on only 3 classes: person, location, organisation.

\subsection{Sentiment Analysis}

With using the manually annotated 5,420 tweet training set to fine-tune a pre-trained multilingual BERT \citep{devlin-etal-2019-bert} model for the sentiment analysis task along with $\sim$20,000 sentiment-annotated Latvian tweets from other sources\footnote{Other Latvian twitter sentiment corpora - \url{https://github.com/Usprogis/Latvian-Twitter-Eater-Corpus/tree/master/sub-corpora/sentiment-analysis\#other-latvian-twitter-sentiment-corpora}}, the 744 tweet test set provided in LTEC can reach an accuracy of around 74\%. 
The accuracy of the model according to the majority of human evaluators on our timeline-balanced 500 tweet set was even higher, reaching 86.40\%. Since our annotation was performed by 12 annotators who did not always agree, we also compared each of their annotations against the majority vote. The accuracy of the average human evaluator compared to the majority was 80.25\%, which does not necessarily indicate human parity of the model, but rather that the correct choice is not always easy to decide.

\section{Conclusion}

In this paper, we introduced an extended version of evaluation sets for the Latvian Twitter Eater Corpus by supplementing the current sentiment analysis test set with named entity labels and full translations into English. We also manually labelled two new evaluation sets for understanding relations that attached images have with the posted text, as well as a more timeline-balanced sentiment analysis test set. 

By experimenting with evaluating pre-trained models using our new test sets we find that some tasks like named entity recognition and sentiment analysis can already handle such data fairly well. However, other tasks like machine translation and especially image-text relation classification still leaves much room for improvement. 

We plan to publish the newly annotated data and merge them into the main repository of LTEC\footnote{Latvian Twitter Eater Corpus - \url{https://github.com/Usprogis/Latvian-Twitter-Eater-Corpus/}}, keeping everything under the current  MIT license.

\section*{Limitations}

In this work, we only considered highly domain-specific data written in a relatively less-spoken language, and possibly containing a noticeable degree of internet slang. Therefore, any results obtained from analysing such data need to be interpreted with these considerations in mind and cannot be generalised to the broader scope of the Latvian language. Also, since hyper-parameter tuning on training large models is computationally very costly, we opt for choosing mostly default parameters and base versions of open-source models for our experiments.

\section*{Ethics Statement}
Our work fully complies with the ACL Code of Ethics\footnote{\url{https://www.aclweb.org/portal/content/acl-code-ethics}}. We use only publicly available datasets and relatively low compute amounts while conducting our experiments to enable reproducibility. All human annotators were fairly compensated for their efforts.

\section*{Acknowledgments}

This paper is based on results obtained from a project JPNP20006, commissioned by the New Energy and Industrial Technology Development Organization (NEDO).

\section{Bibliographical References}\label{sec:reference}

\bibliographystyle{lrec-coling2024-natbib}
\bibliography{lrec-coling,anthology}

\section{Language Resource References}
\label{lr:ref}
\bibliographystylelanguageresource{lrec-coling2024-natbib}
\bibliographylanguageresource{languageresource}

\end{document}